\documentclass[10pt,twocolumn,letterpaper]{article}

\usepackage{iccv}
\usepackage{times}
\usepackage{epsfig}
\usepackage{graphicx}
\usepackage{amsmath}
\usepackage{amssymb}
\usepackage{amsthm}
\usepackage{float}

\usepackage{subfigure}
\usepackage[skip=10pt]{caption}

\usepackage[utf8]{inputenc}
\usepackage[T1]{fontenc}

\usepackage{mathrsfs}
\usepackage[lined,boxed,commentsnumbered]{algorithm2e}
\usepackage{soul}
\usepackage{kbordermatrix}
\usepackage{booktabs}
\usepackage[font=small]{caption}

\usepackage[pagebackref=false,breaklinks=true,letterpaper=true,colorlinks,bookmarks=false]{hyperref}

\newcommand{\mysubsub}[1]{\vspace{0.4em}\noindent\textbf{#1}}
\newcommand{\mysection}[1]{\vspace{-0.15em}\section{#1}\vspace{-0.2em}}
\newcommand{\mysubsection}[1]{\vspace{-0.15em}\subsection{#1}\vspace{-0.2em}}

\newcommand{\R}{\mathbb{R}}
\newcommand{\Y}{\mathcal{Y}}

\newcommand{\Tr}{{\rm Tr}}
\newcommand{\Ybar}{{\rm \overline{\mathcal{Y}}}}

\DeclareMathOperator*{\minprog}{min}

\iccvfinalcopy
\def\iccvPaperID{150} 

\begin{document}

\title{Weakly-Supervised Alignment of Video With Text}

\author{
P. Bojanowski\textsuperscript{1,}\thanks{WILLOW project-team, D\'epartement d'Informatique de l'Ecole Normale Sup\'erieure, ENS/INRIA/CNRS UMR 8548, Paris, France.}
\;\;\;
R. Lajugie\textsuperscript{1,}\thanks{SIERRA project-team, D\'epartement d'Informatique de l'Ecole Normale Sup\'erieure, ENS/INRIA/CNRS UMR 8548, Paris, France.}
\;\;\;
E. Grave\textsuperscript{2,}\thanks{Department of Applied Physics \& Applied Mathematics, Columbia University, New York, NY, USA.} 
\;\;\;
F. Bach\textsuperscript{1,}\footnotemark[2]
\;\;\;
I. Laptev\textsuperscript{1,}\footnotemark[1]
\;\;\;
J. Ponce\textsuperscript{3,}\footnotemark[1]
\;\;\;
C. Schmid\textsuperscript{1,}\thanks{LEAR project-team, INRIA Grenoble Rh\^one-Alpes, Laboratoire Jean Kuntzmann, CNRS, Univ. Grenoble Alpes, France}
\vspace{.4em}\\
\large{\textsuperscript{1}INRIA \quad \textsuperscript{2}Columbia University \quad \textsuperscript{3}ENS / PSL Research University}
\vspace{.4em}\\
}

\maketitle

\begin{abstract}
    Suppose that we are given a set of videos, along with natural language descriptions in the form of multiple sentences (e.g., manual annotations, movie scripts, sport summaries etc.), and that these sentences appear in the same temporal order as their visual counterparts.
    We propose in this paper a method for aligning the two modalities, i.e., automatically providing a time (frame) stamp for every sentence. 
    Given vectorial features for both video and text, this can be cast as a temporal assignment problem, with an implicit linear mapping between the two feature modalities.
    We formulate this problem as an integer quadratic program, and solve its continuous convex relaxation using an efficient conditional gradient algorithm.
    Several rounding procedures are proposed to construct the final integer solution.
    After demonstrating significant improvements over the state of the art on the related task of aligning video with symbolic labels~\cite{Bojanowski14weakly}, we  evaluate our method on a challenging dataset of videos with associated textual descriptions~\cite{Regneri13grounding}, and explore bag-of-words and continuous representations for text.
\end{abstract}

\mysection{Introduction}

Fully supervised approaches to action categorization have shown good performance in short video clips~\cite{Wang13action}.
However, when the goal is not only to classify a clip where a single action happens, but to compute the temporal extent of an action in a long video where multiple activities may take place, new difficulties arise.
In fact, the task of identifying short clips where a single action occurs is at least as difficult as classifying the corresponding action afterwards.
This is reminiscent of the gap in difficulty between categorization and detection in still images.
In addition, as noted in~\cite{Bojanowski14weakly}, manual annotations are very expensive to get, even more so when working with a long video clip or a film shot, where many actions can occur. 
Finally, as mentioned in~\cite{Gaidon11actom,Satkin10a}, it is difficult to define exactly when an action occurs.
This makes the task of understanding human activities much more difficult than finding objects or people in images.

\begin{figure}[t]
    \centering
    \includegraphics[width=\linewidth]{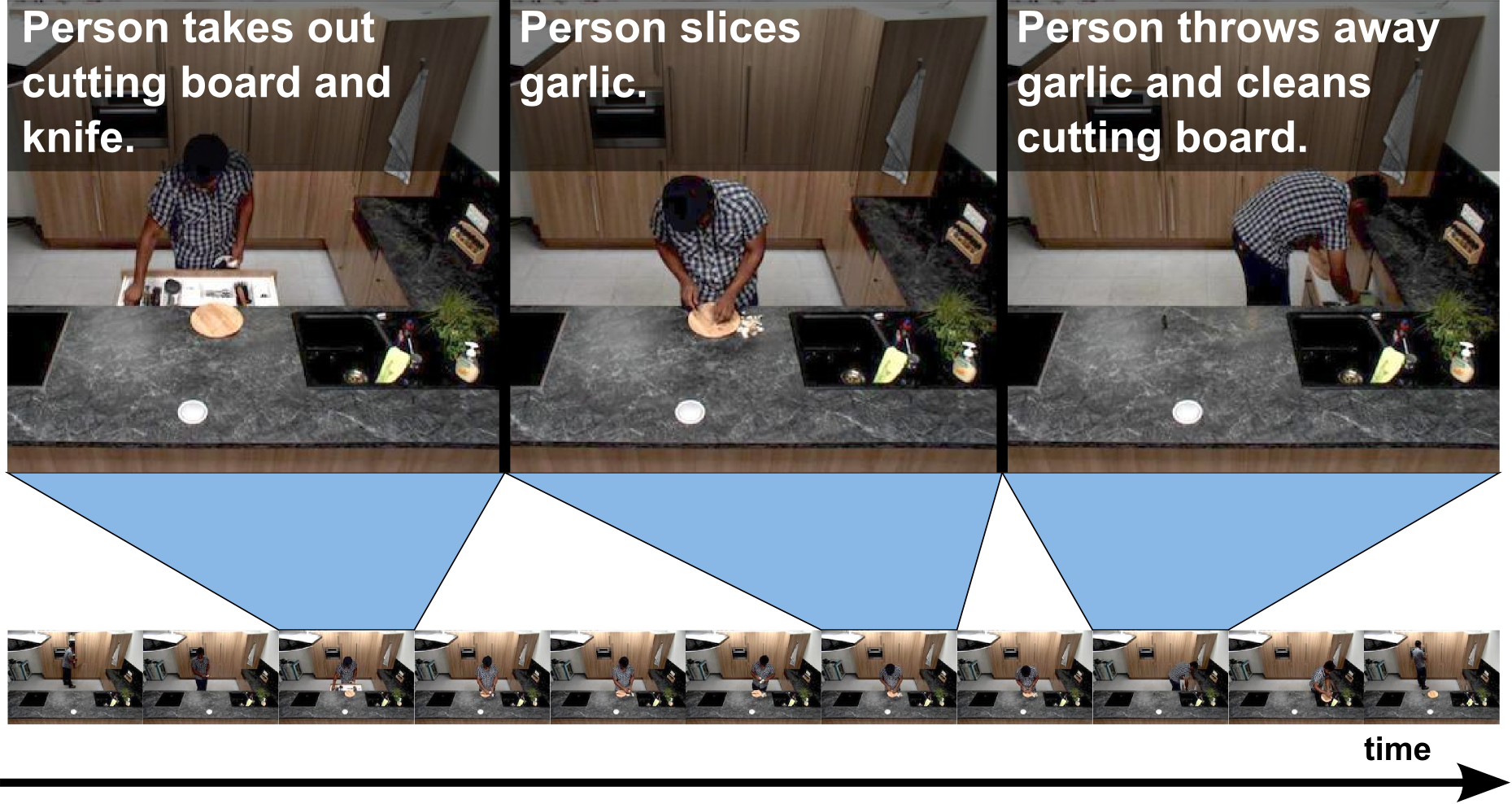}
    \vspace{-2em}
    \caption{
        An example of video to natural text alignment using our method on the TACoS~\cite{Regneri13grounding} dataset.
    }
    \label{fig:teaser}
    \vspace{-2em}
\end{figure}

In this paper, we propose to learn models of video content with minimal manual intervention, using natural language sentences as a weak form of supervision.
This has the additional advantage of replacing purely symbolic and essentially meaningless hand-picked action labels with a semantic representation. 
Given vectorial features for both video and text, we address the problem of temporally aligning the video frames and the sentences, assuming the order is preserved, with an implicit linear mapping between the two feature modalities (Fig.~\ref{fig:teaser}).
We formulate this problem as an integer quadratic program, and solve its continuous convex relaxation using an efficient conditional gradient algorithm.


\mysubsub{Related work.}
Many attempts at automatic image captioning have been proposed over the last decade: Duygulu \etal~\cite{Duygulu02object} were among the first to attack this problem; they proposed to frame image recognition as machine translation.
These ideas were further developed in~\cite{Barnard03matching}.
A second important line of work has built simple natural language models as conditional random fields of a fixed size~\cite{Farhadi10every}.
Typically this corresponds to fixed language templates such as: $\langle$Object, Action, Scene$\rangle$.
Much of the work on joint representations of text and images makes use of canonical correlation analysis (CCA)~\cite{Hotelling1936relations}. 
This approach has first been used to perform image retrieval based on text queries by Hardoon \etal~\cite{Hardoon04canonical}, who learn a kernelized version of CCA to rank images given text.
It has been extended to semi-supervised scenarios~\cite{Socher10connecting}, as well as to the multi-view setting~\cite{Gong14multi}.
All these methods frame the problem of image captioning as a retrieval task~\cite{Hodosh13framing,Ordonez11im2text}. 
Recently, there has also been an important amount of work on joint models for images and text using deep learning (\eg~\cite{Frome13devise,Karpathy14deepFragment,Lebret15phrase,Socher14grounded}).

There has been much less work on joint representations for text and video.
A dataset of cooking videos with associated textual descriptions is used to learn joint representations of those two modalities in~\cite{Regneri13grounding}.
The problem of video description is framed as a machine translation problem in~\cite{Rohrbach13translating}, while a deep model for descriptions is proposed in~\cite{Donahue14long}.
Recently, a joint model of text, video and speech has also been proposed~\cite{Malmaud15what}.
Textual data such as scripts, has been used for automatic video understanding, for example for action recognition~\cite{Laptev08a,Marszalek09a}.
Subtitles and scripts have also often been used to guide person recognition models (\eg~\cite{Bojanowski13finding,Ramanathan14linking,Tapaswi12knock}).

The temporal structure of videos and scripts has been used in several papers. 
In~\cite{Bojanowski14weakly}, an action label is associated with every temporal interval of the video while respecting the order given by some annotations (see~\cite{Ramanathan14linking} for related work). 
The problem of aligning a large text corpus with video is addressed in~\cite{Tapaswi15book}.
The authors propose to match a book with its television adaptation by solving an alignment problem.
This problem is however very different from ours, since the alignment is based only on character identities.
The temporal ordering of actions, \eg, in the form of Markov models or action grammars, has been used to constrain action prediction in videos~\cite{Kwak11scenario,Laxton07leveraging,Ryoo06recognition}.
Spatial and temporal constraints have also been used in the context of group activity recognition~\cite{Amer13monte,Khamis12combining}.
Similarily to our work, \cite{Zhou09canonical} uses a quadratic objective under time warping constraints. 
However it does not provide a convex relaxation, and proposes an alternate optimization method instead. 
Time warping problems under constraints have been studied in other vision tasks, especially to address the challenges of large scale data~\cite{Rakthanmanon2013addressing}.

The model we propose in this work is based on discriminative clustering, a weakly supervised framework for partitioning data. 
Contrary to standard clustering techniques, it uses a discriminative cost function~\cite{Bach07diffrac,Guo2007convex} and it has been used in image co-segmentation~\cite{Joulin10discriminative,Joulin12a}, object co-localization~\cite{Tang2014efficient}, person identification in video~\cite{Bojanowski13finding,Ramanathan14linking}, and alignment of labels to videos~\cite{Bojanowski14weakly}.
Contrary to~\cite{Bojanowski14weakly}, for example, our work makes use of continuous text representations.
Vectorial models for words are very convenient when working with heterogeneous data sources.
Simple sentence representations such as bags of words are still frequently used~\cite{Gong14multi}.
More complex word and sentence representations can also be considered.
Simple models trained on a huge corpus~\cite{Mikolov13distributed} have demonstrated their ability to encode useful information.
It is also possible to use different embeddings, such as the posterior distribution over latent classes given by a hidden Markov model trained on the text~\cite{Grave14markovian}.

\mysubsection{Problem statement and approach}
\begin{figure}[t]
    \centering
    \includegraphics[width=0.8\linewidth]{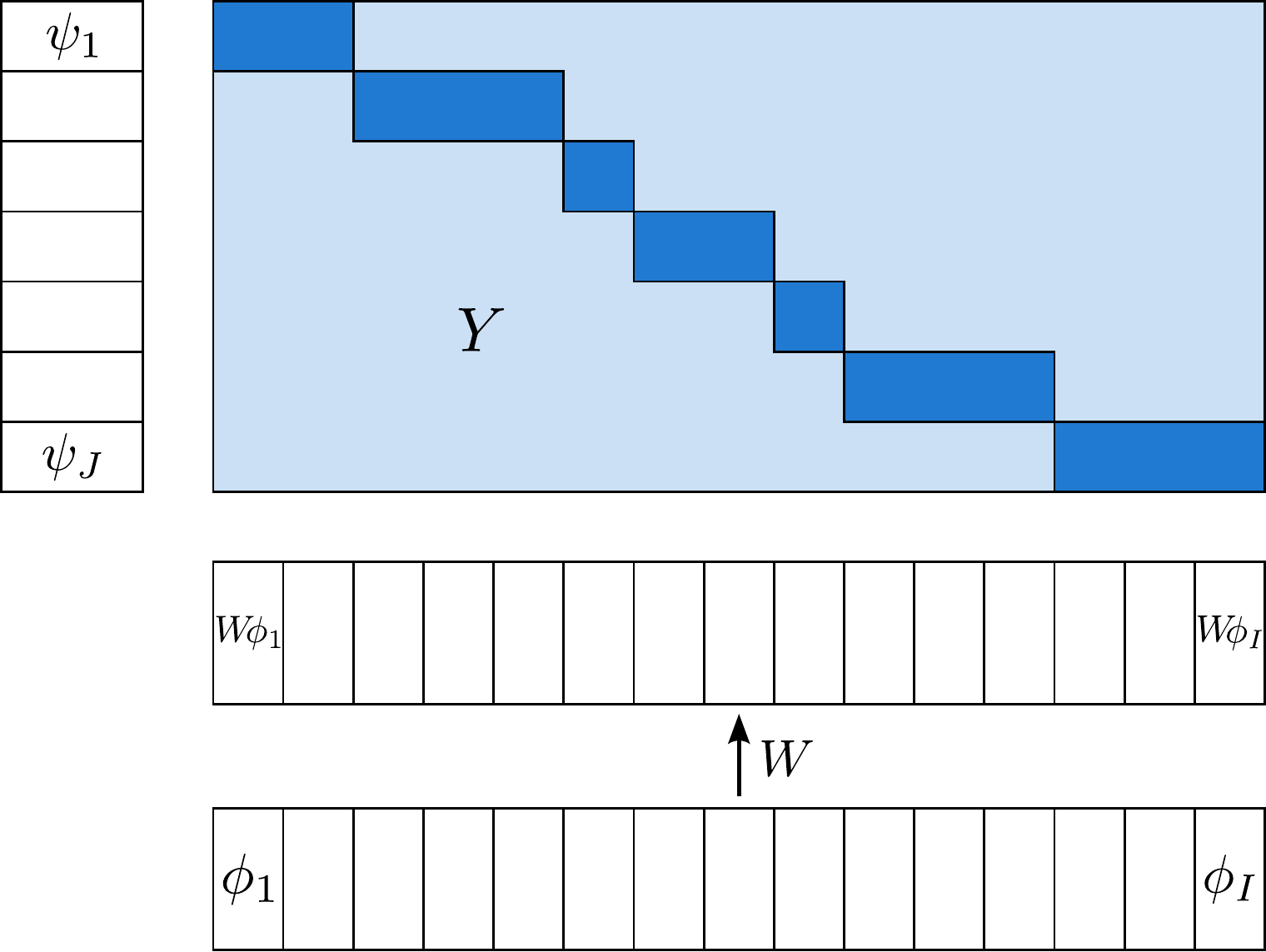}
    \caption{
        Illustration of some of the notations used in this paper.
        The video features $\Phi$ are mapped to the same space  as text features using the map $W$.
        The temporal alignment of video and text features is encoded by the assignment matrix $Y$.
        Light blue entries in $Y$ are zeros, dark blue entries are ones.
        See text for more details.\label{fig:notations}
    }
    \vspace{-1.5em}
\end{figure}
\mysubsub{Notation.}
Let us assume that we are given a data stream, associated with two modalities, represented by the features $\Phi = [\phi_1, \dots, \phi_I]$ in $\mathbb{R}^{D \times I}$ and \(\Psi = [\psi_1, \dots, \psi_J]\) in $\mathbb{R}^{E \times J}$.
In the context of video to text alignment, $\Phi$ is a description of the video signal, made up of $I$ temporal intervals, and $\Psi$ is a textual description, composed of $J$ sentences. 
However, our model is general and can be applied to other types of sequential data (biology, speech, music, \etc).
In the rest of the paper, except of course in the experimental section, we stick to the abstract problem, considering two generic modalities of a data stream.

\mysubsub{Problem statement.}
Our goal is to assign every element $i$ in $\{1, \dots, I\}$ to exactly one element $j$ in $\{1, \dots, J\}$. 
At the same time, we also want to learn a linear map\footnote{As usual, we actually want an affine map. This can be done by simply adding a constant row to $\Phi$.} between the two feature spaces, parametrized by \(W\) in $\mathbb{R}^{E \times D}$.
If the element~$i$ is assigned to an element~$j$, we want to find $W$ such that $\psi_j \approx W \phi_i$.
If we encode the assignments in a binary matrix~$Y$, this can be written in matrix form as: $\Psi Y  \approx W \Phi$ (Fig.~\ref{fig:notations}).
The precise definition of the matrix $Y$ will be provided in Sec.~\ref{sec:notations}.
In practice, we insert zero vectors in between the columns of $\Psi$. 
This allows some video frames not to be assigned to any text.

\mysubsub{Relation with Bojanowski \etal~\cite{Bojanowski14weakly}.}
Our model is an extension of~\cite{Bojanowski14weakly} with several important improvements.
In~\cite{Bojanowski14weakly}, instead of aligning video with natural language, the goal is to align video to symbolic labels in some predefined dictionary of size $K$ (``open door'', ``sit down'', \etc). 
By representing the labeling of the video using a matrix $Z$ in $\{0,1\}^{K \times I}$, the problem solved there corresponds to finding $W$ and $Z$ such that: $Z \approx W \Phi$.
The matrix $Z$ encodes both data (which labels appear in each clip and which order) and the actual temporal assignments.
Our parametrization allows us instead to separate the representation $\Psi$ from the assignment variable $Y$.
This has several significant advantages:
first, this allows us to consider continuous text representations as the predicted output \(\Psi\) in $\R^{E \times J}$ instead of just classes.
As shown in the sequel, this also allows us to easily impose natural, data-independent constraints on the assignment matrix $Y$.

\mysubsub{Contributions.}
This article makes three main contributions:
\textbf{(i)} we extend the model proposed in~\cite{Bojanowski14weakly} in order to work with continuous representations of text instead of symbolic classes;
\textbf{(ii)} we propose a simple method for including prior knowledge about the assignment into the model; and
\textbf{(iii)} we demonstrate the performance of the proposed model on challenging video datasets equipped with natural language meta data.

\mysection{Proposed model}
\label{sec:notations}

\mysubsection{Basic model}
Let us begin by defining the binary \emph{assignment matrices} \(Y\) in \(\{0, 1\}^{J \times I}\).
The entry \(Y_{ji}\) is equal to one if \(i\) is assigned to \(j\) and zero otherwise.
Since every element $i$ is assigned to exactly one element $j$, we have that \(Y^T \mathbf{1}_{J} = \mathbf{1}_{I}\), where $\mathbf{1}_k$ represents the vector of ones in dimension $k$.
As in~\cite{Bojanowski14weakly}, we assume that temporal ordering is preserved in the assignment.
Therefore, if the element $i$ is assigned to $j$, then $i+1$ can only be assigned to $j$ or $j+1$.
In the following, we will denote by \(\Y\) the set of matrices \(Y\) that satisfy this property. 
Our recursive definition allows us to obtain an efficient dynamic programming algorithm for minimizing linear functions over $\Y$, which is a key step to our optimization method.

We measure the discrepancy between $\Psi Y$ and $W \Phi$ using the squared $L_2$ loss.
Using an $L_2$ regularizer for the model $W$, our learning problem can now be written as:
\vspace{-0.4em}
\begin{equation}
\label{eq:discriminativeAlone}
\minprog_{
    Y \in \mathcal{Y}
} 
\
\min_{
    W \in \mathbb{R}^{E \times D}
}
 \ \frac{1}{2I} \|\Psi Y - W \Phi \|_F^2 + \frac{\lambda}{2}\|W\|_F^2.
\end{equation}
We can rewrite~(\ref{eq:discriminativeAlone}) as: \(\minprog_{Y \in \Y} \  q(Y)\), where $q: \Y \to \mathbb{R}$ is defined for all \(Y\) in \(\Y\) by:
\begin{equation}
    q(Y) = 
    \min_{
            W \in \mathbb{R}^{H \times D} 
    } 
    \left [ 
        \frac{1}{2I} \|\Psi Y - W \Phi \|_F^2
        + \frac{\lambda}{2} \|W\|_F^2
    \right ].
    \label{eq:ridge}
\end{equation}
For a fixed \(Y\), the minimization with respect to \(W\) in~(\ref{eq:ridge}) is a ridge regression problem. 
It can be solved in closed form, and its solution is:
\begin{align}
    W^* &= \Psi Y \Phi^T \left ( \Phi \Phi^T + I \lambda \text{Id}_D \right )^{-1},
\end{align}
where $\text{Id}_k$ is the identity matrix in dimension $k$.
Substituting in~(\ref{eq:ridge}) yields:
\begin{equation}
    \label{eq:q}
    q(Y) = \frac{1}{2I} \text{Tr} 
    \left ( 
        \Psi Y Q Y^T \Psi^T
    \right ),
\end{equation}
where $Q$ is a matrix depending on the data and the regularization parameter $\lambda$:
\begin{equation}
    Q =  
        \text{Id}_I - \Phi^T 
        \left (
            \Phi \Phi^T + I \lambda \text{Id}_D
            \right )^{-1}
        \Phi
    .
\end{equation}

\mysubsub{Multiple streams.}
Suppose now that we are given $N$ data streams (videos in our case), indexed by $n$ in $\{1, \dots, N\}$.
The approach proposed so far is easily generalized to this case by taking $\Psi$ and $\Phi$ to be the horizontal concatenation of all the matrices $\Psi_n$ and $\Phi_n$. 
The matrices $Y$ in $\Y$ are block-diagonal in this case, the diagonal blocks being the assignment matrices of every stream: 
\[
    Y = 
    \begin{bmatrix}
        Y_1 &  & 0 \\
            & \ddots & \\
          0 &  & Y_N 
    \end{bmatrix}.
\]
This is the model actually used in our implementation.

\mysubsection{Priors and constraints}
\label{sec:priors-and-constraints}
We can incorporate task-specific knowledge in our model by adding constraints on the matrix $Y$ to model event duration for example.
Constraints on $Y$ can also be used to avoid the degenerate solutions known to plague discriminative clustering~\cite{Bach07diffrac,Bojanowski14weakly,Guo2007convex,Joulin10discriminative}.

\mysubsub{Duration priors.}
The model presented so far is solely based on a discriminative function.
Our formulation in terms of an assignment variable \(Y\) allows us to reason about the number of elements $i$ that get assigned to the element $j$. For videos, since each element $i$ correponds to a fixed time interval, this number is the \emph{duration} of text element $j$.
More formally, the duration \(\delta(j)\) of element $j$ is obtained as: $\delta(j) = \mathbf{e}_j^T Y \mathbf{1}_{I}$, 
where \(\mathbf{e}_j\) is the \(j\)-th vector of the canonical basis of \(\mathbb{R}^{J}\).
Assuming for simplicity a single target duration \(\mu\) and variance parameter \(\sigma\) for all units, this leads to the following duration penalty:
\begin{equation}
    r(Y) = \frac{1}{2 \sigma^2} \|Y \mathbf{1}_{I} - \mu \|_2^2.
    \label{eq:duration}
\end{equation}

\begin{figure}[t]
    \centering
    \subfigure[A (near) degenerate solution.]{
        \includegraphics[width=0.45\linewidth]{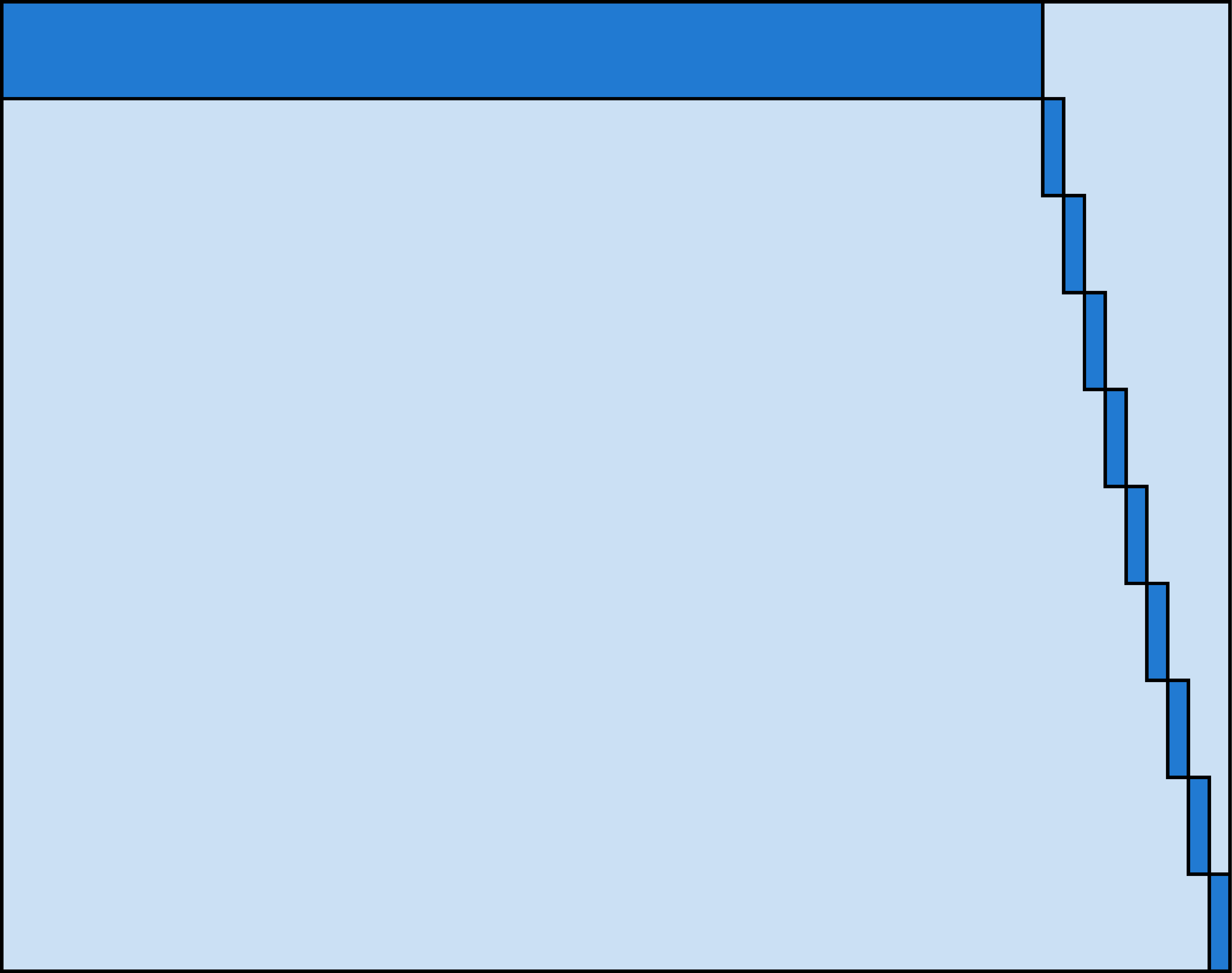}
    }
    \
    \subfigure[A constrained solution.]{
            \includegraphics[width=0.45\linewidth]{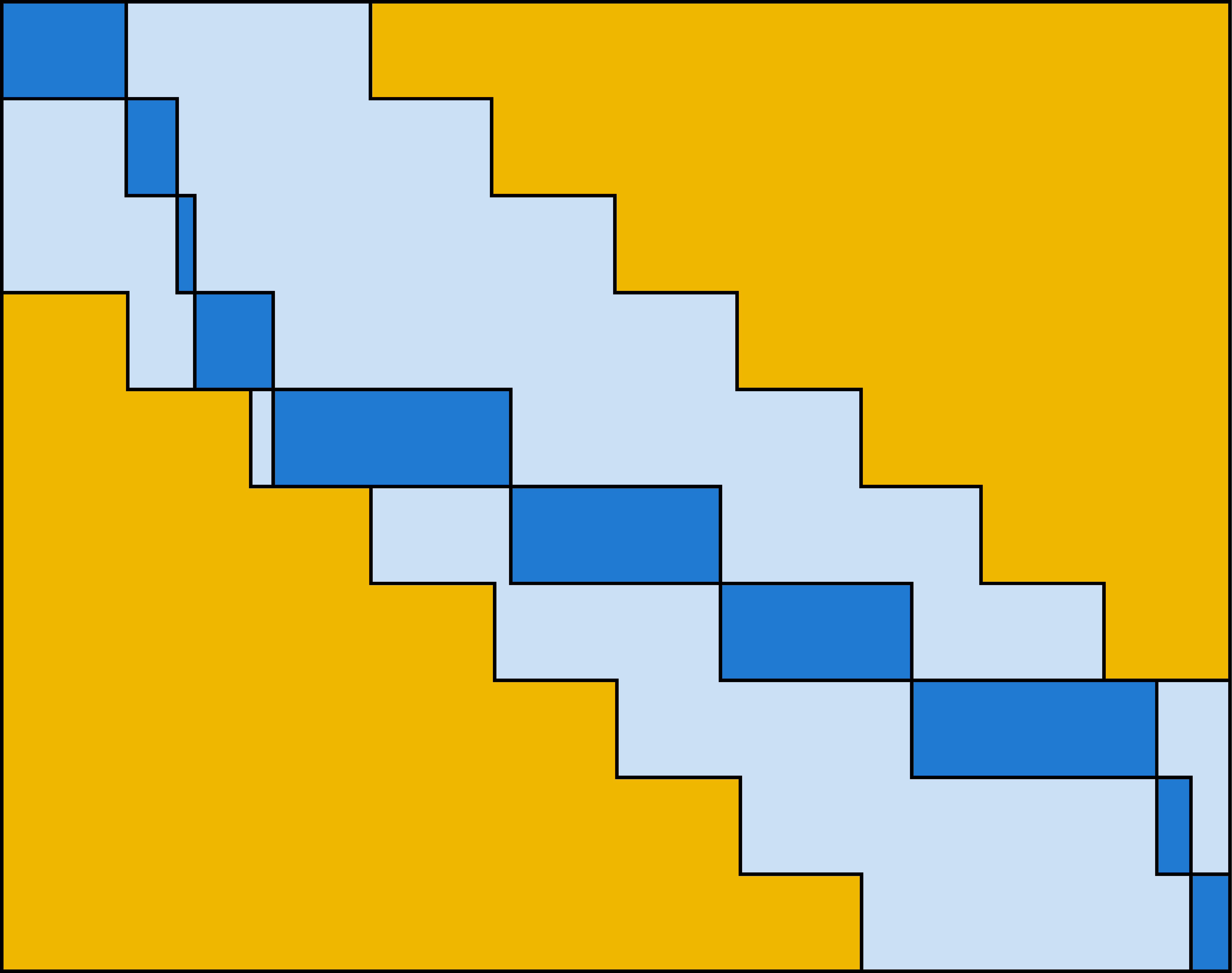}
    }
    \vspace{-1em}
    \caption{
        \textbf{(a)} depicts a typical near degenerate solution where almost all the the elements $i$ are assigned to the first element, close to the constant vector element of the kernel of $Q$. 
        \textbf{(b)} We propose to avoid such solutions by forcing the alignment to stay outside of a given region (shown in yellow), which may be a band or a parallelogram.
        The dark blue entries correspond to the assignment matrix $Y$, and the yellow ones represent the constraint set.
        See text for more details.
        (Best seen in color.)
    }
    \label{fig:dp-constraints}
    \vspace{-1em}
\end{figure}

\mysubsub{Path priors.}
\label{sec:bands}
Some elements of $\Y$ correspond to very unlikely assignments.
In speech processing and various related tasks~\cite{Rabiner93fundamentals}, the warping paths are often constrained, forcing for example the path to fall in the Sakoe-Chiba band or in the Itakura parallelogram~\cite{sakoe78dynamic}.
Such constraints allow us to encode task-specific assumptions and to avoid degenerate solutions associated with the fact that constant vectors belong to the kernel of $Q$ (Fig.~\ref{fig:dp-constraints}~(a)).
Band constraints, as illustrated in Fig.~\ref{fig:dp-constraints}~(b), successfully exclude the kind of degenerate solutions presented in (a).
Let us denote by \(Y_c\) the band-diagonal matrix of width $\beta$, such that the diagonal entries are $0$ and the others are $1$; such a matrix is illustrated in Fig.~\ref{fig:dp-constraints}~(b) in yellow.
In order to ensure that the assignment does not deviate too much from the diagonal, we can impose that at most $C$ non zero entries of $Y$ are outside the band.
We can formulate that constraint as follows: $\text{Tr} (Y_c^T Y) \leq C$.

This constraint could be added to the definition of the set $\Y$, but this would prohibit the use of dynamic programming, which is a key step to our optimization algorithm described in Sec.~\ref{sec:opt}.
We instead propose to add a penalization term to our cost function, corresponding to the Lagrange multiplier for this constraint.
Indeed, for any value of $C$, there exists an $\alpha$ such that if we add
\begin{equation}
    \vspace{-0.2em}
    l(Y) = \alpha \text{Tr}(Y_c^T Y) ,
    \label{eq:band}
    \vspace{-0.2em}
\end{equation}
to our cost function, the two solutions are equal, and thus the constraint is satisfied.
In practice, we select the value of $\alpha$ by doing a grid search on a validation set.

\mysubsection{Full problem formulation}
Including the constraints defined in Sec.~\ref{sec:priors-and-constraints} into our objective function yields the following optimization problem:
    \vspace{-1em}
\begin{align}
    \label{eq:full-prob}
\minprog_{
        Y \in \Y 
} 
\ q(Y) + r(Y) + l(Y),
\end{align}
where $q$, $r$ and $l$ are the three functions respectively defined in~(\ref{eq:q}),~(\ref{eq:duration}) and~(\ref{eq:band}).




\mysection{Optimization}
\label{sec:opt}

\mysubsection{Continuous relaxation}
The discrete optimization problem formulated in Eq.~(\ref{eq:full-prob}) is the minimization of a positive semi-definite quadratic function over a very large set $\Y$, composed of binary assignment matrices.
Following~\cite{Bojanowski14weakly}, we relax this problem by minimizing our objective function over the (continuous) convex hull~$\Ybar$ instead of~$\Y$.
Although it is possible to describe $\Ybar$ in terms of linear inequalities, we never use this formulation in the following, since the use of a general linear programing solver does not exploit the structure of the problem. 
Instead, we consider the relaxed problem:
\vspace{-0.49em}
\begin{equation}
    \label{eq:probY-relaxed}
    \minprog_{Y \in \Ybar} \ q(Y) + r(Y) + l(Y)
\end{equation}
as the minimization of a convex quadratic function over an implicitly defined convex and compact domain.
This type of problem can be solved efficiently using the Frank-Wolfe algorithm~\cite{Bojanowski14weakly,Frank1956} as soon as it is possible to minimize linear forms over the convex compact domain.

First, note that $\Ybar$ is the convex hull of $Y$, and the solution to $\minprog_{Y \in \Y}\Tr(AY)$ is also a solution of $\minprog_{Y \in \Ybar}\Tr(AY)$~\cite{Bertsekas99nonlinear}.
As noted in~\cite{Bojanowski14weakly}, it is possible to minimize any linear form $\Tr(AY)$, where $A$ is an arbitrary matrix, over $\Y$ using dynamic programming in two steps: 
First, we build the cumulative cost of matrix $D$ whose entry $(i,j)$ is the cost of the optimal alignment starting in $(1,1)$ and terminating in $(i,j)$. 
This step can be done recursively in $\mathcal{O} (IJ)$ steps. 
Second, we recover the optimal $Y$ by backtracking in the matrix $D$. See~\cite{Bojanowski14weakly} for details.

\mysubsection{Rounding}
\label{sec:rounding}
Solving~(\ref{eq:probY-relaxed}) provides a continuous solution $Y^*$ in $\Ybar$ and a corresponding optimal linear map $W^*$.
Our original problem is defined on \(\Y\), and we thus need to round $Y^*$. 
We propose three rounding procedures, two of them corresponding to Euclidean norm minimization and a third one using the map $W^*$.
All three roundings boil down to solving a linear problem over $\Y$, which can be done once again using dynamic programming.
Since there is no principled, analytical way to pick one of these procedures over the others, we conduct an empirical evaluation in Sec.~\ref{sec:experimental} to assess their strengths and weaknesses.

\mysubsub{Rounding in $\Y$.}
The simplest way to round $Y^*$ is to find the closest point $Y$  according to the Euclidean distance in the space $\Y$: $\minprog_{Y \in \Y} \|Y - Y^*\|_F^2$.
This problem can be reduced to a linear program over $\Y$.

\mysubsub{Rounding in $\Psi \Y$.}
This is in fact the space where the original least-squares minimization is formulated.
We solve in this case the problem $\minprog_{Y \in \Y} \|\Psi (Y - Y^*)\|_F^2$, which weighs the error measure using the feature \(\Psi\).
A simple calculation shows that the previous problem is equivalent to:
\begin{equation}
    \minprog_{Y \in \Y} \ \text{Tr} 
    \left ( 
        Y^T 
        \left ( 
        \mathbf{1}_I \text{Diag}(\Psi^T \Psi)^T - 2 \Psi^T \Psi Y^*
        \right ) 
    \right ).
\end{equation}

\mysubsub{Rounding in $W$.}
Our optimization procedure gives us two outputs, namely a relaxed assignment $Y^* \in \overline{\Y}$ and a model $W^*$ in $\R^{E \times D}$. 
We can use this model to predict an alignment $Y$ in $\Y$ by solving the following quadratic optimization problem: $\minprog_{Y \in \Y} \|\Psi Y - W^*\Phi\|_F^2$.
As before, this is equivalent to a linear program. 
An important feature of this rounding procedure is that it can also be used on previously unseen data.


\mysection{Semi-supervised setting}

\begin{figure}[t]
    \centering
    \subfigure[$Y$ fixed to ground truth.]{
        \includegraphics[width=0.45\linewidth]{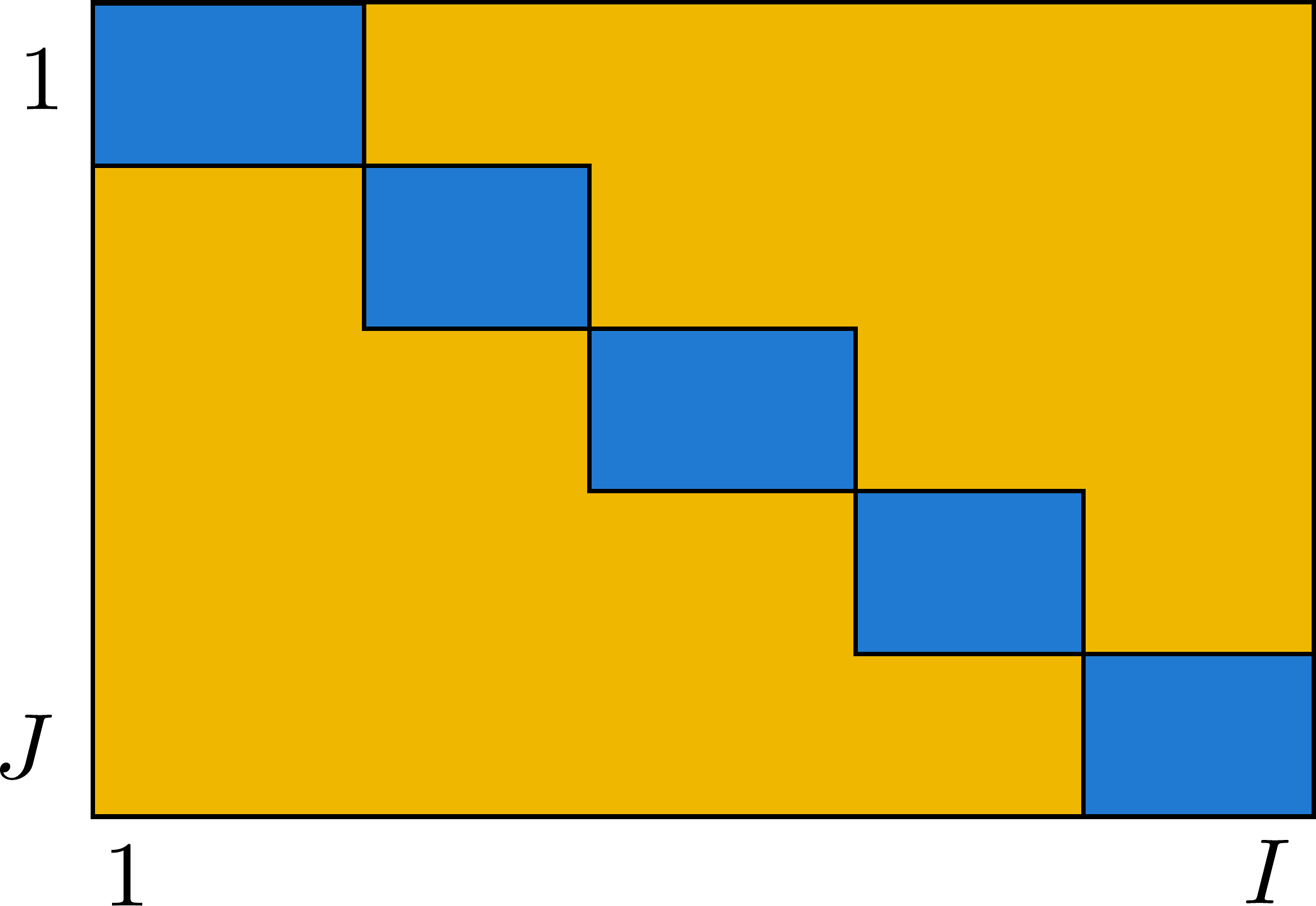}
    }
    \subfigure[Corresponding constraints.]{
        \includegraphics[width=0.45\linewidth]{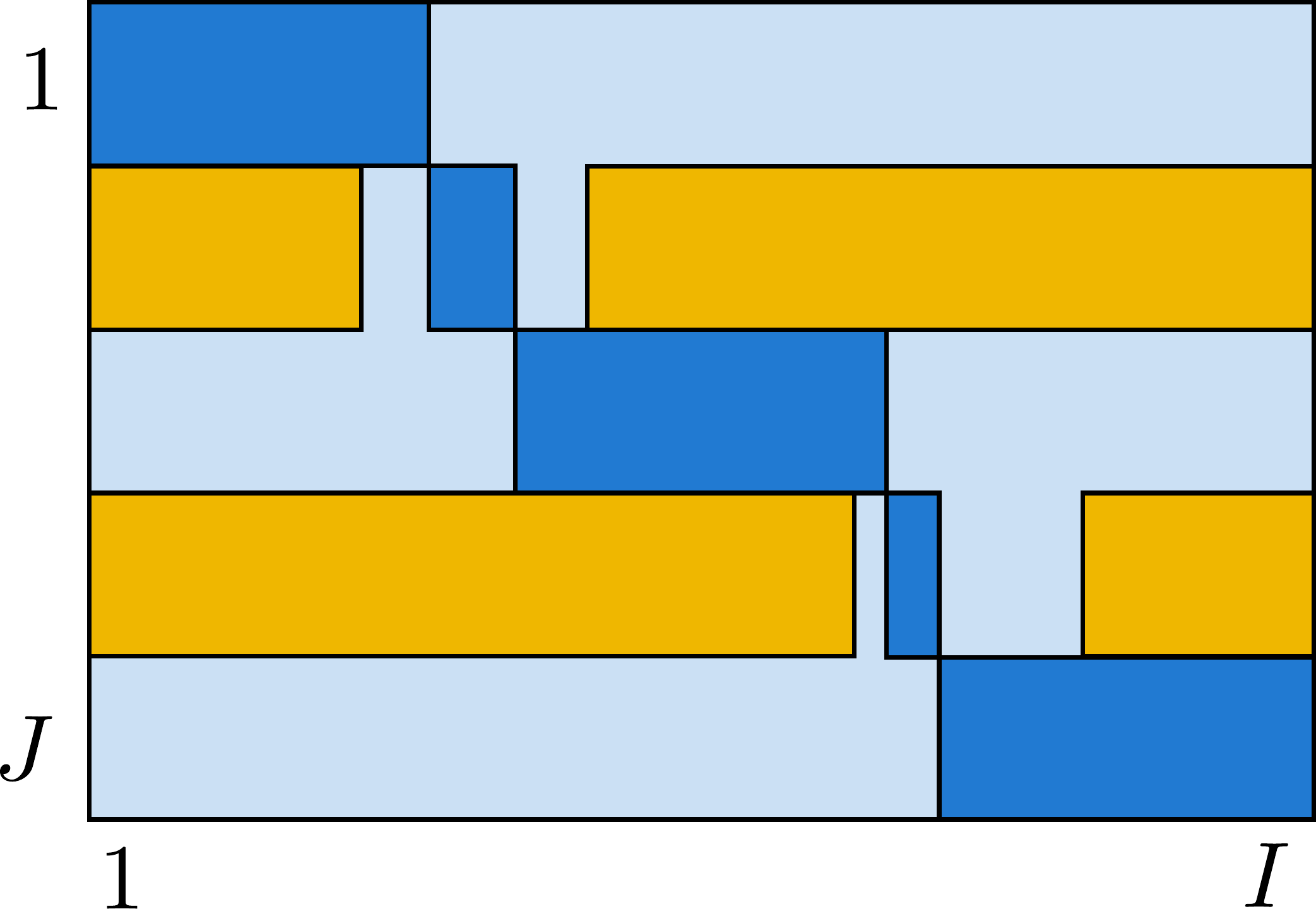}
    }
    \vspace{-1em}
    \caption{
        Two ways of incorporating supervision.
        \textbf{(a)} the assignments are fixed to the ground truth:
        the dark blue entries exactly correspond to $Y_s$, and yellow entries are forbidden assignments;
        \textbf{(b)} the assignments are constrained. 
        For even rows, assignments must be outside the yellow strips.
	    Light blue regions correspond to authorized paths for the assignment. 
    }
    \label{fig:semisup}
    \vspace{-0.5em}
\end{figure}

The proposed model is well suited to semi-supervised learning.
Incorporating additional supervision just consists in constraining parts of the matrix \(Y\).
Let us assume that we are given a triplet \((\Psi_s, \Phi_s, Y_s)\) representing supervisory data.
The part of data that is not involved in that supervision is denoted by \((\Psi_u, \Phi_u, Y_u)\).
Using the additional data amounts to solving~\eqref{eq:full-prob} with matrices \((\Psi, \Phi, Y)\) defined as:
\begin{equation}
    \Psi = [\Psi_u, \ \kappa \ \Psi_s],
    \Phi = [\Phi_u, \ \kappa \ \Phi_s],
    Y = 
    \begin{bmatrix}
        Y_u & 0 \\
          0 & Y_s
    \end{bmatrix}.
\end{equation}
The parameter \(\kappa\) allows us to weigh properly the supervised and unsupervised examples.
Scaling the features this way corresponds to using the following loss:
\begin{align}
     \|\Psi_u Y_u - W \Phi_u \|_F^2 + \kappa^2  \|\Psi_s Y_s - W \Phi_s\|_F^2. 
\end{align}
Since $Y_s$ is given, we can optimize over $\Y$ while constraining the lower right block of $Y$.
In our implementation this means that we fix the lower-right entries in $Y$ to the ground-truth values during optimization.

Manual annotations of videos are sometimes imprecise, and we thus propose to include them in a softer manner.
As mentioned in Sec.~\ref{sec:notations}, odd columns in $\Psi$ are filled with zeros.
This allows some video frames not to be assigned to any text.
Instead of imposing that the assignment \(Y\) coincides with the annotations, we constrain it to lie within annotated intervals.
For any even (non null) element $j$, we force the set of video frames that are assigned to $j$ to be a subset of those in the ground truth (Fig.~\ref{fig:semisup}). 
That way, we allow the assignment to pick the most discriminative parts of the video within the annotated interval.
This way of incorporating supervision empirically yields much better performance.


\mysection{Experimental evaluation}\label{sec:experimental}

We evaluate the proposed approach on two challenging datasets.
We first compare it to a recent method on the associated dataset~\cite{Bojanowski14weakly}.
We then run experiments on TACoS, a video dataset composed of cooking activities with textual annotations~\cite{Regneri13grounding}.
We select the hyper parameters $\lambda, \alpha, \sigma, \kappa$ on a validation set.
All results are reported with standard error over several random splits.

\mysubsub{Performance measure.}
All experiments are evaluated using the \emph{Jaccard measure} in~\cite{Bojanowski14weakly}, that quantifies the difference between a ground-truth assignment $Y_{gt}$ and the predicted $Y$ by computing the precision for each row. 
In particular the best performance of $1$ is obtained if the predicted assignment is within the ground-truth.
If the prediction is outside, it is equal to $0$. 

\mysubsection{Comparison with Bojanowski et al.~\cite{Bojanowski14weakly}}
Our model is a generalization of Bojanowski \etal~\cite{Bojanowski14weakly}.
Indeed, we can easily cast the problem formulated in that paper into our framework.
Our model differs from the aforementioned one in three crucial ways:
First, we do not need to add a separate ``background class'', which is always problematic.
Second, we propose another way to handle the semi-supervised setting. 
Most importantly, we replace the matrix $Z$ by $\Psi Y$, allowing us to add data-independent constraints and priors on $Y$.
In this section we describe comparative experiments conducted on the dataset proposed in~\cite{Bojanowski14weakly}.

\mysubsub{Dataset.}
We use the videos, labels and features provided in~\cite{Bojanowski14weakly}. 
This data is composed of 843 videos (94 videos are set aside for a classification experiement) that are annotated with a sequence of labels.
There are 16 different labels such as \eg ``Eat'', ``Open Door'' and ``Stand Up''.
As in the original paper, we randomly split the dataset into ten different validation, evaluation and supervised sets.

\mysubsub{Features.}
The label sequences provided as weak supervisory signal in~\cite{Bojanowski14weakly} can be used as our features \(\Psi\).
We consider a language composed of sixteen words, where every word corresponds to a label.
Then, the representation \(\psi_j\) of every element \(j\) is the indicator vector of the $j$-th label in the sequence.
Since we do not model background, we simply interleave zero vectors in between meaningful elements.
The matrix \(\Phi\) corresponds to the video features provided with the paper's code.
These features are 2000-dimensional bag-of-words vectors computed on the HOF channel.

\mysubsub{Baselines.}
As our baseline, we run the code from~\cite{Bojanowski14weakly} that is available online\footnote{https://github.com/piotr-bojanowski/action-ordering} for different fractions of annotated data, seeds and parameters.
As a sanity check, we compare the performance of our algorithm to that of a random assignment that follows the priors.
This random baseline obtains a performance measure of $32.8$ with a standard error of $0.3$.

\begin{figure}[t]
    \centering
    \includegraphics[width=0.75\linewidth]{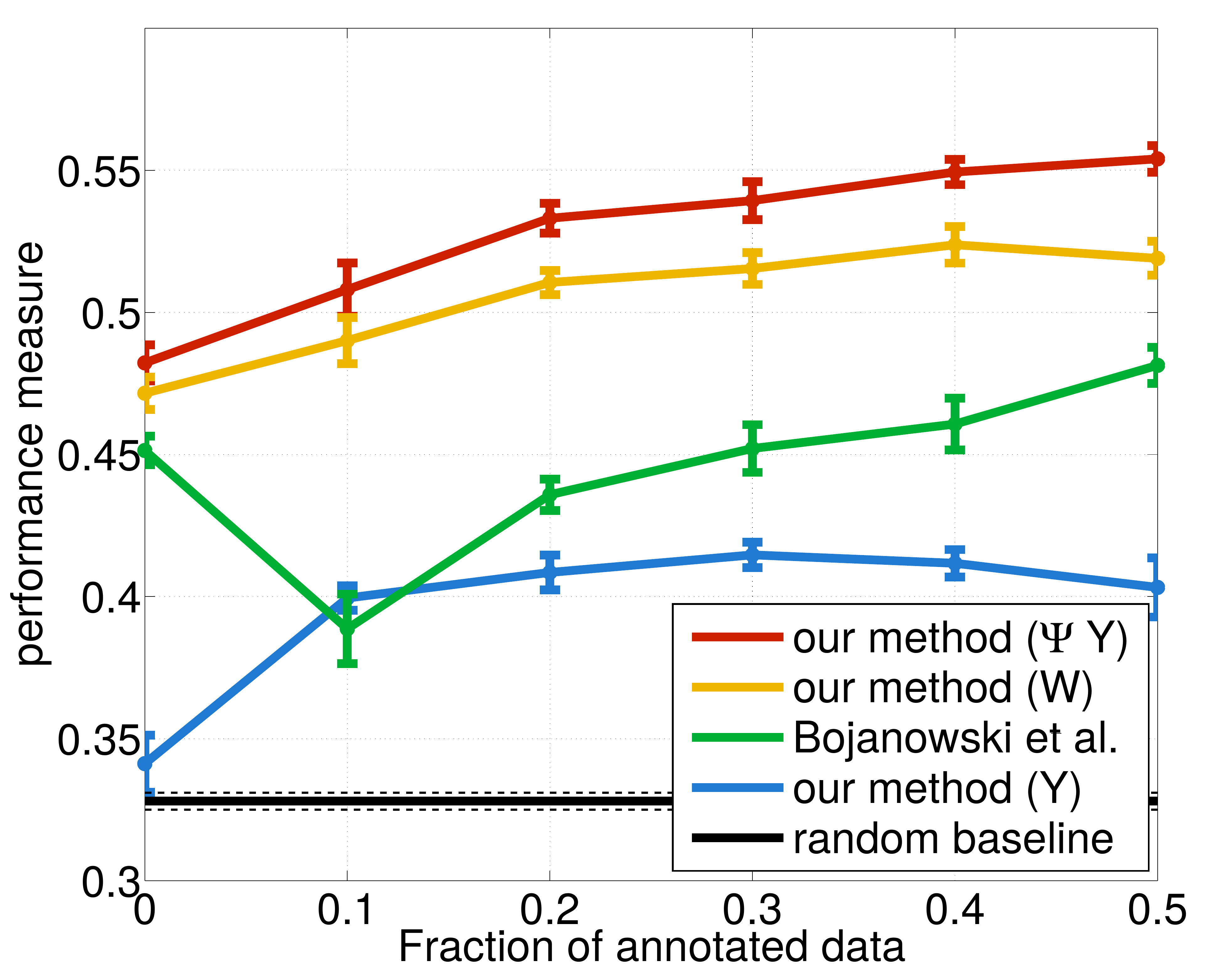}
    \vspace{-0.7em}
    \caption{
        Comparing our approach with the various rounding schemes to the model in~\cite{Bojanowski14weakly} on the same data, using  the same evaluation metric as in~\cite{Bojanowski14weakly}. See text for details.
    }
    \label{fig:eccv}
    \vspace{-1.5em}
\end{figure}

\mysubsub{Results.}
We plot performance versus amount of supervised data in Fig.~\ref{fig:eccv}.
We use the same evaluation metric as in~\cite{Bojanowski14weakly}.
First of all, when no supervision is available, our method works significantly better (no overlap between error bars).
This can be due (1) to the fact that we do not model background as a separate class;
and (2) to the use of the priors described in Sec.~\ref{sec:priors-and-constraints}.
As additional supervisory data becomes available, we observe a consistent improvement of more than 5\% over~\cite{Bojanowski14weakly} for the $\Psi Y$ and $W$ roundings.
The $Y$ rounding does not give good results in general. 

The main interesting point is the fact that the drop at the beginning of the curve in~\cite{Bojanowski14weakly} does not occur in our case.
When no supervised data is available, the optimal $Y^*$ solely depends on the video features $\Phi$.
When the fraction of annotated data increases, the optimal $Y^*$ changes and depends on the annotations.
However, the temporal extent of an action is not well defined.
Therefore, manual annotations need not be coherent with the $Y^*$ obtained with no supervision.
Our way of dealing with supervised data is less coercive and does not commit strictly to the annotated data.

In Fig.~\ref{fig:eccv} we have observed that the best performing rounding on this task is the one using the matrix product $\Psi Y$.
It is important to notice that~\cite{Bojanowski14weakly} performs rounding on a matrix $Z=\Psi Y$ which is thus equivalent to the best performing rounding for our method.
In preliminary experiments, we observed that using a $W$ rounding for~\cite{Bojanowski14weakly} does not significantly improve performance.

\mysubsection{Results on the TACoS dataset}
We also evaluate our method on the TACoS dataset~\cite{Regneri13grounding} which includes actual natural language sentences. 
On this dataset, we use the $W$ rounding as it is the one that empirically gives the best test performance.
We do not have yet a compelling explanation as to why this is the case.

\mysubsub{Dataset.}
TACoS is composed of 127 videos picturing people who perform cooking tasks.
Every video is associated with two kinds of annotations.
The first one is composed of low-level activity labels with precise temporal location.
We do not make use of these fine-grained annotations in this work.
The second one is a set of natural language descriptions that were obtained by crowd-sourcing.
Annotators were asked to describe the content of the video using simple sentences.
\begin{figure*}[th]
    \centering
    \subfigure[duration prior]{
        \includegraphics[width=0.28\linewidth]{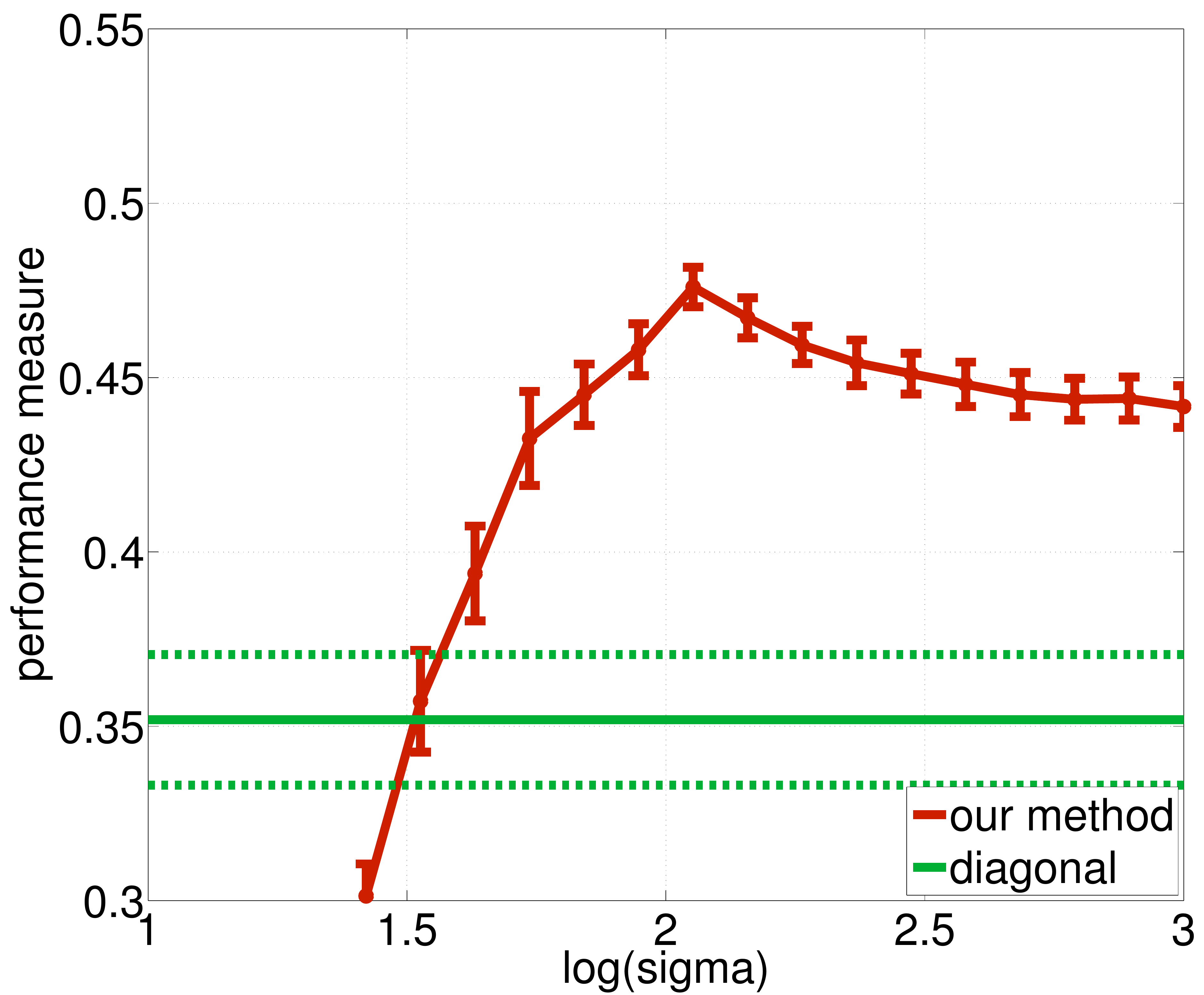}
    }
    \subfigure[band prior]{
        \includegraphics[width=0.28\linewidth]{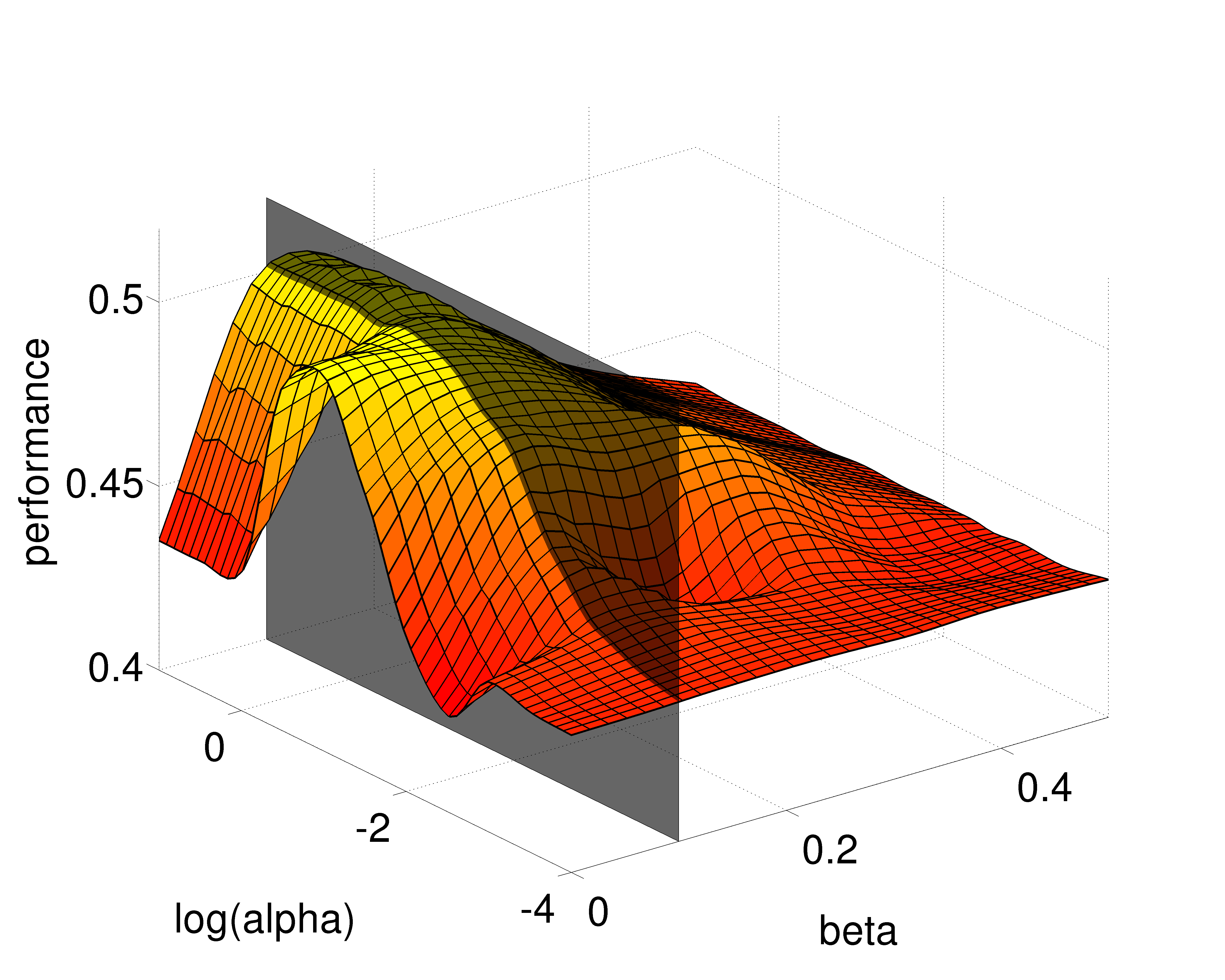}
    }
    \subfigure[band prior]{
        \includegraphics[width=0.28\linewidth]{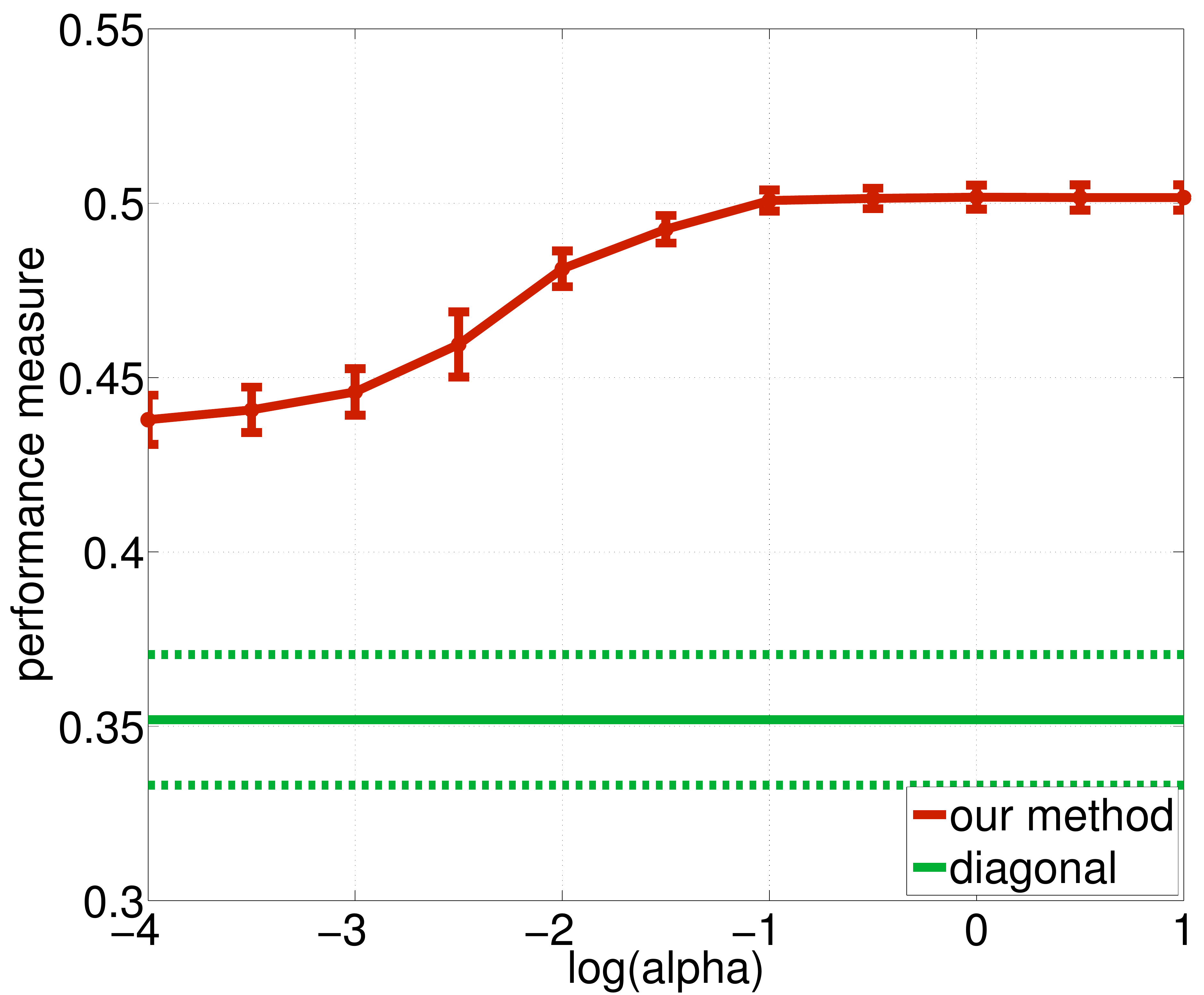}
    }
    \vspace{-1em}
    \caption{
        Evaluation of the priors we propose in this paper.
        \textbf{(a)} We plot the performance of our model for various values of $\sigma$.
        When $\sigma$ is big, the prior has no effect.
        We see that there is a clear trade off and an optimal choice of $\sigma$ yields better performance.
        \textbf{(b)} Performance as a function of $\alpha$ and width of the band.
        The shown surface is interpolated to ease readability.
        \textbf{(c)} Performance for various values of $\alpha$.
        This plot corresponds to the slice illustrated in (b) by the black plane.
    }
    \label{fig:priors}
    \vspace{-0.48cm}
\end{figure*}
Each video $\Phi$ is associated with $k$ textual descriptions $[\Psi^1, \dots, \Psi^K]$.
Every textual description is composed of multiple sentences with associated temporal extent.
We consider as data points the pairs $(\Psi^k, \Phi)$ for $k$ in $\{1, \dots, K\}$.

\mysubsub{Video features.}
We build the feature matrix $\Phi$ by computing dense trajectories~\cite{Wang13action} on all videos.
We compute dictionaries of 500 words for HOG, HOF and MBH channels.
These experimentally provide satisfactory performance while staying relatively low-dimensional.
For a given temporal window, we concatenate bag-of-words representations for the four channels.
As in the Hellinger kernel, we use the square root of $L_1$ normalized histograms as our final features.
We use overlapping temporal windows of length 150 frames with a stride of 50.

\mysubsub{Text features.}
To apply our method to textual data, we need a feature representation $\psi_i$ for each sentence. 
In our experiments, we explore multiple ways to represent sentences and empirically compare their performance (Table~\ref{tab:textreps}).
We discuss two ways to encode sentences into vector representations, one based on bag of words, the other on continuous word embeddings~\cite{Mikolov13distributed}.

To build our bag-of-words representation, we construct a dictionary using all sentences in the TACoS dataset.
We run a part-of-speech tagger and a dependency parser~\cite{Manning14core} in order to exploit the grammatical structure.
These features are pooled using three different schemes.
(1) ROOT: In this setup, we simply encode each sentence by its root verb as provided by the dependency parser. 
(2) ROOT+DOBJ: In this setup we encode a sentence by its root verb and its direct object dependency.
This representation makes sense on the TACoS dataset as sentences are in general pretty simple. 
For example, the sentence ``The man slices the cucumber'' is represented by ``slice'' and ``cucumber''.
(3) VNA: This representation is the closest to the usual bag-of-words text representation.
We simply pool all the tokens whose part of speech is Verb, Noun or Adjective.
The two first representations are very rudimentary versions of bags of words.
They typically contain only one or two non zero elements.

We also explore the use of word embeddings (W2V)~\cite{Mikolov13distributed}, trained on three different corpora.
First, we train them on the TACoS corpus.
Even though the amount of data is very small (175,617 words), the vocabulary is also limited and the sentences are simple. 
Second, we train the vector representations on a dataset of 50,993 kitchen recipes, downloaded from allrecipes.com.
This corresponds to a corpus of roughly 5 million tokens.
However, the sentences are written in imperative mode, which differs from the sentences found in TACoS.
For completeness, we also use the WaCky corpus~\cite{Baroni09wacky}, a large web-crawled dataset of news articles.
We train representations of dimension 100 and 200.
A sentence is then represented by the concatenation of the vector representations of its root verb and its root's direct object.

\begin{table}[b]
    \centering
    \vspace{-1em}
    \begin{tabular}{lcc}
        \toprule
        text representation   & Dim. 100 & Dim. 200 \\
        \midrule
        W2V UKWAC                & 43.8 (1.5) & 46.4 (0.7) \\
        W2V TACoS                & 48.3 (0.4) & 48.2 (0.4)\\
        W2V ALLRECIPE           & 43.3 (0.7) & 44.7 (0.5)\\
        \bottomrule
    \end{tabular}
    \caption{
      Comparison of text representations trained on different corpora, in dimension $100$ and $200$.
    }
    \label{tab:textreps}
\end{table}

\mysubsub{Baselines.}
On this dataset, we considered two baselines.
The first one is Bojanowski \etal~\cite{Bojanowski14weakly} using the ROOT textual features. 
Verbs are used in place of labels by the method.
The second one, that we call Diagonal, corresponds to the performance obtained by the uniform alignment, \ie assigning the same amount of video elements $i$ to each textual element $j$.

%


\begin{figure*}[th]
    \centering
    \includegraphics[width=\linewidth]{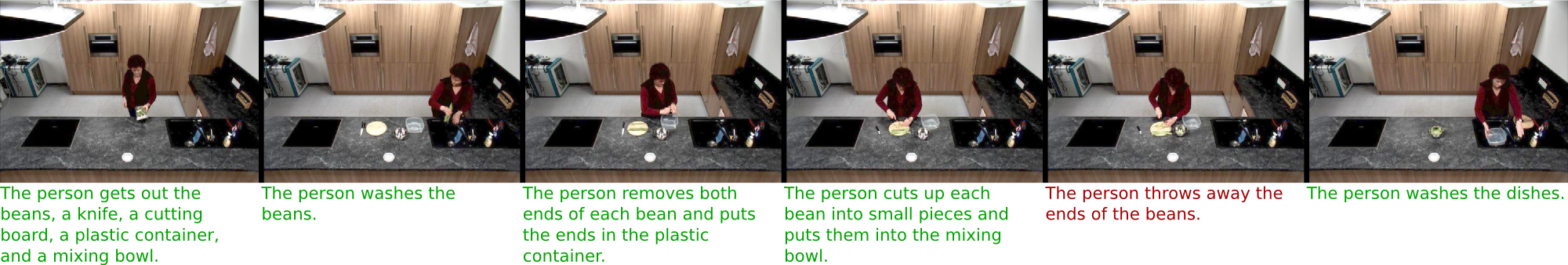}\\\vspace{0.3em}
    \includegraphics[width=\linewidth]{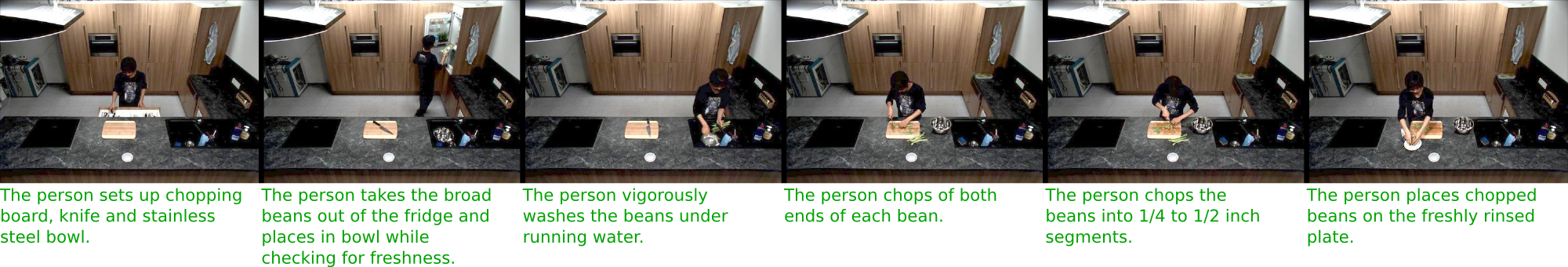}\\\vspace{0.3em}
    \includegraphics[width=\linewidth]{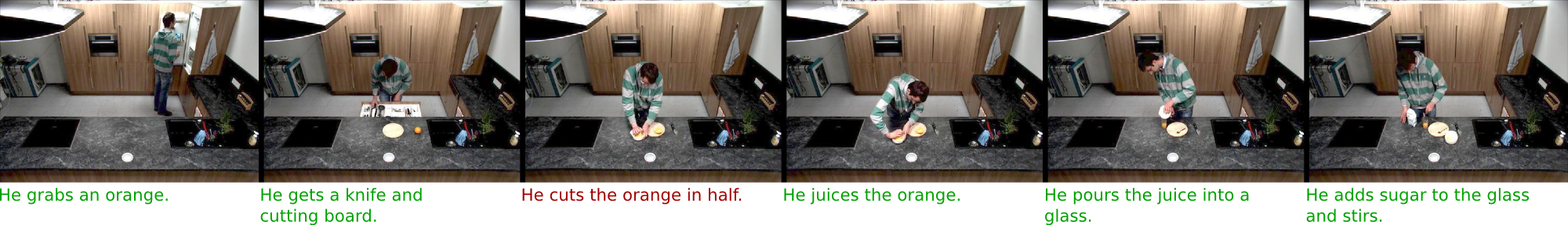}
    \vspace{-1.5em}
    \caption{
        Representative qualitative results for our method applied on TACoS.
        Correctly assigned frames are in green, incorect ones in red.
    }
    \label{fig:qualitative}
    \vspace{-1.5em}
\end{figure*}

\mysubsub{Evaluation of the priors.}
We proposed in Sec.~\ref{sec:priors-and-constraints} two heuristics for including prior knowledge and avoiding degenerate solutions to our problem.
In this section, we evaluate the performance of these priors on TACoS.
To this end, we run our method with the two different models separately.
We perform this experiment using the ROOT+DOBJ text representation.
The results of this experiment are illustrated in Fig.~\ref{fig:priors}.

We see that both priors are useful.
The duration prior, when $\sigma$ is carefully chosen, allows us to improve performance from $0.441$ (infinite $\sigma$) to $0.475$.
There is a clear trade-off in this parameter.
Using a bit of duration prior helps us to get a meaningful $Y^*$ by discarding degenerate solutions.
However, when the prior is too strong, we obtain a degenerate solution with decreased performance.

\begin{table}[b]
    \centering
    \vspace{-1em}
    \begin{tabular}{lll}
        \toprule
        text representation   & nosup & semisup \\
        \midrule
              Diagonal              & \multicolumn{2}{c}{35.2 (3.7)} \\
              \textit{Bojanowski et al.} \cite{Bojanowski14weakly}   & 39.0 (1.0) & 49.1 (0.7)\\ 
        \midrule
        ROOT                        & 49.9 (0.2) & 59.2 (1.0)\\
        ROOT+DOBJ                   & 48.7 (0.9) & 65.4 (1.0)\\
        VNA                         & 45.7 (1.4) & 59.9 (2.9)\\
        W2V TACoS 100               & 48.3 (0.4) & 60.2 (1.5)\\
        \bottomrule
    \end{tabular}
    \caption{
        Performance when no supervision is used to compute the assignment (nosup) and when half of the dataset is provided with time stamped sentences (semisup).
    }
    \label{tab:semisup}
\end{table}

The band prior (as depicted in Fig.~\ref{fig:priors}, b and c) improves performance even more.
We plot in (b) the performance as a joint function of the parameter $\alpha$ and of the width of the band $\beta$.
We see that the width that provides the best performance is 0.1.
We plot in (c) the corresponding performance as a function of $\alpha$.
Using large values of $\alpha$ corresponds to constraining the path to be entirely inside the band, which explain why the performance flattens for large $\alpha$.
When using a small width, the best path is not entirely inside the band and one has to carefully choose the parameter $\alpha$.

We show in Fig.~\ref{fig:priors} the performance of our method for various values of the parameters on the evaluation set.
Please note however that when used in other experiments, the actual values of these parameters are chosen on the validation set only.
Sample qualitative results are shown in Fig.~\ref{fig:qualitative}


\mysubsub{Evaluation of the text representations.}
In Table~\ref{tab:textreps}, we compare the continuous word representations trained on various text corpora.
The representation trained on TACoS works best.
It is usually advised to retrain the representation on a text corpus that has similar distribution to the corpus of interest.  
Moreover, higher-dimensional representations (200) do not help probably because of the limited vocabulary size.
The representations trained on a very large news corpus (UKWAC) benefits from using higher-dimensional vectors.
With such a big corpus, the representations of the cooking lexical field are probably merged together.
This is further demonstrated by the fact that using embedings trained on Google News provided weak performance (42.1).

In Table~\ref{tab:semisup}, we experimentally compare our approach to the baselines, in an unsupervised setting and a semi-supervised one.
First, we observe that the diagonal baseline has reasonable performance. 
Note that this diagonal assignment is different from a random one since a uniform assignment between text and video in our context makes some sense. 
Second, we compare to the method of~\cite{Bojanowski14weakly} on ROOT, which is the only set up where this method can be used.
This baseline is higher than the diagonal one but pretty far from the performances of our model using ROOT as well.

Using bag-of-words representations, we notice that simple pooling schemes work best.
The best performing representation is purely based on verbs.
This is probably due to the fact that richer representations can mislead such a weakly supervised method.
As additional supervision becomes available, the ROOT+DOBJ pooling works much better that only using ROOT validating the previous claim.

\vspace{-0.5em}
\mysection{Discussion.}
We presented in this paper a method able to align a video with its natural language description.
We would like to extend our work to even more challenging scenarios including feature movies and more complicated grammatical structures.
Also, our use of natural language processing tools is limited, and we plan to incorporate better grammatical reasoning in future work.

\mysubsub{Acknowledgements.}
\small{
    This work was supported in part by a PhD fellowship from the EADS Foundation, the Institut Universitaire de France and ERC grants Activia, Allegro, Sierra and VideoWorld. 
}

{
\vspace{-0.5em}
\small
\bibliographystyle{ieee}
\bibliography{biblio}
}

\end{document}